\documentclass[11pt,a4paper]{article}
\usepackage{authblk}
\usepackage[hyperref]{ranlp2023}
\usepackage{times}
\usepackage{latexsym}

\usepackage{microtype}

\usepackage{amsfonts}
\usepackage{amsmath}
\usepackage{multirow}
\usepackage{adjustbox}
\usepackage{booktabs}
\usepackage{arydshln}
\usepackage{titlesec}
\usepackage{soul}
\usepackage{subfig}
\usepackage{dirtytalk}

\aclfinalcopy 
\def\aclpaperid{35} %  Enter the acl Paper ID here

\makeatletter
\def\adl@drawiv#1#2#3{%
        \hskip.5\tabcolsep
        \xleaders#3{#2.5\@tempdimb #1{1}#2.5\@tempdimb}%
                #2\z@ plus1fil minus1fil\relax
        \hskip.5\tabcolsep}
\newcommand{\cdashlinelr}[1]{%
  \noalign{\vskip\aboverulesep
           \global\let\@dashdrawstore\adl@draw
           \global\let\adl@draw\adl@drawiv}
  \cdashline{#1}
  \noalign{\global\let\adl@draw\@dashdrawstore
           \vskip\belowrulesep}}
\makeatother

\newcommand{\quotes}[1]{``#1''}
\renewcommand{\vec}[1]{\ensuremath{\mathbf{#1}}}

\usepackage{fancyhdr}

\fancypagestyle{firstpage}{
  \fancyhf{} % clear all fields for first page
  \fancyfoot[C]{\thepage} % page number centered
  \fancyfoot[L]{\scriptsize Accepted at RANLP 2023. Official version: \href{https://aclanthology.org/2023.ranlp-1.118/}{aclanthology.org/2023.ranlp-1.118/}} % left side footer text
}

\title{Prompt-Based Approach for Czech Sentiment Analysis}

\author[*]{\bf Jakub \v{S}m\'{i}d}
\author[*,$\dagger$]{\bf  Pavel P\v{r}ib\'{a}\v{n}}
\affil[ ]{University of West Bohemia, Faculty of Applied Sciences, Czech Republic}
\affil[*]{Department of Computer Science and Engineering,}
\affil[$\dagger$]{NTIS -- New Technologies for the Information Society,}
\affil[  ]{\tt	\{jaksmid,pribanp\}@kiv.zcu.cz}
\affil[  ]{\tt http://nlp.kiv.zcu.cz}

\date{}
\begin{document}
\maketitle

\thispagestyle{firstpage} % use custom footer on first page

\begin{abstract}
This paper introduces the first prompt-based methods for aspect-based sentiment analysis and sentiment classification in Czech. We employ the sequence-to-sequence models to solve the aspect-based tasks simultaneously and demonstrate the superiority of our prompt-based approach over traditional fine-tuning. In addition, we conduct zero-shot and few-shot learning experiments for sentiment classification and show that prompting yields significantly better results with limited training examples compared to traditional fine-tuning. We also demonstrate that pre-training on data from the target domain can lead to significant improvements in a zero-shot scenario.
\end{abstract}

% ============================== Introduction =====================================
\section{Introduction}
\label{sec:introduction}
\par In recent years, pre-trained BERT-like \citep{devlin-etal-2019-bert} models based on the Transformer \citep{attention-all-transformer} architecture and language modelling significantly improved the performance of various NLP tasks \citep{raffel2020exploring-t5}. The initial approach was to pre-train these models on a large amount of text and then fine-tune them for a specific task. More recently, an approach exploiting the nature of language modelling appeared, called \textit{prompting} or \textit{prompt-based fine-tuning}. Prompting is a technique that encourages a pre-trained model to make specific predictions by providing a prompt specifying the task to be done \citep{propmting-overview}.

\par This new approach became very popular in solving NLP problems in zero-shot or few-shot scenarios, including sentiment analysis \citep{gao-etal-2021-making,gao-etal-2022-lego,hosseini-asl-etal-2022-generative}. Most of the current research aimed at languages other than Czech, especially English. To the best of our knowledge, no research has focused on any sentiment analysis task in the Czech language using prompt-based fine-tuning. To address this lack of research, this paper presents an initial study focusing on two sentiment-related tasks: \textbf{aspect-based sentiment analysis} and \textbf{sentiment classification} in the Czech language by applying prompt-based fine-tuning.

\par The \textit{sentiment classification} (SC), also known as \textit{polarity detection}, is a classification task where the objective for a given text is to assign one overall sentiment polarity label. Usually, the three-class scheme with \textit{positive}, \textit{negative} and \textit{neutral} labels is used, but more labels can be applied \cite{liu2012sentiment}.

\par
\begin{figure}[ht!]
    \centering
    \includegraphics[width=7.5cm]{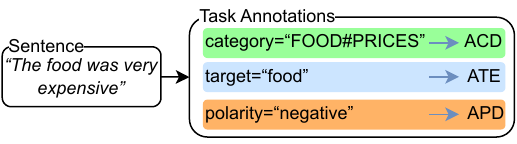}
    \caption{The example of the ABSA tasks.}
    \label{fig:absa-example}
\end{figure}

 Aspect-based sentiment analysis (ABSA) is a more detailed task compared to SC, which aims to extract fine-grained information about entities, their aspects and opinions expressed towards them. Generally, the goal of ABSA is to identify the sentiment of each aspect or feature of a product or service. There are multiple definitions and versions of the ABSA task \citep{pontiki-etal-2014-semeval,saeidi-etal-2016-sentihood,barnes-etal-2022-semeval}. In this work, we focus on the version of \textit{aspect-based sentiment analysis} presented in the SemEval competitions \citep{pontiki-etal-2015-semeval,pontiki-etal-2016-semeval}, which includes several subtasks. Specifically, the tasks are aspect category detection (ACD), aspect term extraction (ATE), simultaneously detecting (aspect category, aspect term) tuples (ACTE), and detecting the sentiment polarity (APD)\footnote{The ACD, ATE, ACTE and APD tasks are named Slot1, Slot2, Slot1\&2 and Slot3, respectively, in \citep{pontiki-etal-2015-semeval,pontiki-etal-2016-semeval} under Subtask 1.}  of a given aspect term and category (see Figure \ref{fig:absa-example} for examples).

\par In addition, we solve the target-aspect-sentiment detection task (TASD) \citep{tasd}, which aims to simultaneously detect the aspect category, aspect term and sentiment polarity.

\par
This paper presents a novel approach for solving Czech sentiment classification and ABSA tasks using prompt-based fine-tuning. We utilize Czech monolingual BERT-like models and their language modelling ability to perform \textit{prompting} for the APD and SC tasks. We use multilingual text-to-text generative models for the remaining ABSA tasks to generate textual predictions based on prompted input. Our approach enables us to solve all these ABSA tasks at once, and we show that it is superior to the traditional fine-tuning approach for them.

\par
We also explore zero-shot and few-shot learning scenarios for APD and SC tasks and show that prompting leads to significantly better results with fewer training examples compared to traditional fine-tuning. Additionally, we demonstrate that pre-training on data from a target domain results in great improvements in a zero-shot scenario.

\par
Our study provides pioneered results for prompt-based fine-tuning in Czech sentiment. Overall, our key contributions are the following: 1) We propose, to the best of our knowledge, the first prompt-based approach for sentiment analysis tasks in Czech. 2) We show the superior performance of our prompting approach over traditional fine-tuning for ABSA tasks. 3) We compare the two approaches and show that prompting achieves better results than traditional fine-tuning in few-shot scenarios.

\section{Related Work}
\par
This section reviews prior works conducted on sentiment analysis in Czech.
The prompt-based fine-tuning is a relatively new paradigm in NLP, and to the best of our knowledge, no research has yet explored its application on sentiment analysis in Czech. To partly address this research gap, we include prompt-based approaches for analogous sentiment analysis tasks in English.

\subsection{Czech Sentiment Classification} 
\par
The first approaches for sentiment analysis in Czech often utilized lexical features \citep{steinberger-etal-2011-creating-trian,veselovska2012creating} and n-gram text representations in combination with classifiers like maximum entropy or Naive Bayes \citep{habernal-etal-2013-sentiment}. Subsequently, \citet{brychcin-habernal-2013-unsupervised-sentiment} employed a mixture of supervised and unsupervised techniques to improve polarity detection in movie reviews. Similarly to \citet{kim-2014-convolutional}, \citet{SLON-Lenc2016Neural} used the convolutional neural network (CNN) and Long Short-Term Memory (LSTM) for SC of the same CSFD dataset we use in this work, see Section \ref{sec:sentiment-classification-data}. The authors of \citep{libovicky2018solving} added self-attention to an LSTM-based neural network and applied it to the CSFD dataset. A detailed survey of older approaches for Czech sentiment analysis is presented by \citet{cano2019sentiment}. 
In recent years, Czech Transformer-based models have been proposed and have shown great success in Czech sentiment analysis. \citet{sido2021czert} introduced the first Czech BERT-like model, outperforming previous state-of-the-art (SotA) results in SC. Additionally, \citet{straka2021robeczech} presented a pre-trained Czech version of the RoBERTa \citep{zhuang-etal-2021-robustly} model and demonstrated its effectiveness for the Czech language on Facebook posts. \citet{priban-steinberger-2021-multilingual} provide the SotA results for three Czech polarity detection datasets.
The most recent work comes from \citet{priban-tsd-2022,priban-steinberger-2022-czech}, where the authors investigate the possibility of performing zero-shot cross-lingual sentiment analysis and subjectivity classification between Czech and English with multilingual Transformer-based models like mBERT \citep{devlin-etal-2019-bert} or XLM-R \citep{xlm-r}.

\subsection{Czech Aspect-Based Sentiment Analysis} 
The ABSA task in the Czech language has been much less studied in recent years and the existing approaches are usually outdated compared to recent sentiment classification methods. The pioneering research on Czech ABSA can be found in \citet{steinberger-etal-2014-aspect}, where the authors manually annotated and created a restaurant review dataset for the same tasks as in the SemEval 2014 competition \citep{pontiki-etal-2014-semeval}. They provided results of baseline models based on Conditional Random Fields (CRF) and Maximum Entropy (ME) classifier. \citet{tamchyna2015czech} built a dataset containing IT product reviews and provided baseline results with the CRF. Unlike in the mentioned Czech restaurant dataset, the IT product reviews are annotated with global sentiment and aspect terms but without any categorization and sentiment toward the terms. \citet{hercig2016unsupervised} extended the Czech restaurant review ABSA dataset and suggested several unsupervised methods to enhance the performance on ABSA tasks in Czech and English using the CRF and ME classifiers. They showed that unsupervised methods can provide substantial improvements.

\subsection{Prompt-Based and Related Approaches}
\par As we already mentioned, there is no work for Czech sentiment analysis based on prompt-based fine-tuning. Therefore, we provide example studies focused on English sentiment analysis using prompt-based approaches or related methods.

\par \citet{zhang-etal-2021-towards-generative} formulate the ABSA tasks as a text generation problem. They propose two paradigms to deal with the ABSA tasks, namely annotation-style and extraction-style modelling, both generating textual output in a desired format. They utilize the English T5 \citep{raffel2020exploring-t5} text-to-text Transformer-based model and evaluate their approach on various ABSA tasks, including the TASD task, on datasets from the SemEval competitions \citep{pontiki-etal-2014-semeval,pontiki-etal-2015-semeval,pontiki-etal-2016-semeval}. They showed the effectiveness of their approach by establishing new SotA results. Similarly, \citet{zhang-etal-2021-aspect-sentiment} used the same English T5 model to solve a newly introduced ABSA task called \textit{aspect sentiment quad prediction} by generating textual output. Another approach proposed by \citet{gao-etal-2022-lego} aims to solve multiple ABSA tasks at once. The authors applied the English T5 model to a prompt created from the individual ABSA tasks. They evaluated their model on the same datasets as \citet{zhang-etal-2021-towards-generative}, outperforming the previously mentioned approach and achieving new SotA results.

\par \citet{gao-etal-2021-making} experimented with prompt-based fine-tuning for SC. With the English T5 model, they automatically generated prompts for BERT and RoBERTa models, which they consequently fine-tuned for the SC task. They demonstrated that their few-shot approach leads to better results compared to traditional fine-tuning.

\section{Data \& Tasks Definition}
In this section, we describe the aspect-based and sentiment classification datasets. Furthermore, we describe in more detail the ABSA tasks introduced in Section \ref{sec:introduction}, on which this paper is focused.

\subsection{Data for Sentiment Classification}
\label{sec:sentiment-classification-data}
For the SC task, where the goal is to assign one overall polarity label (\textit{positive}, \textit{negative} or \textit{neutral}) for a given text, we employ the Czech CSFD dataset \citep{habernal-etal-2013-sentiment}. The dataset contains 91,381 movie reviews from the Czech movie database\footnote{\label{foot:github}\url{https://www.csfd.cz}}. The reviews are annotated in a distant supervised way according to the star rating assigned to each review (0–1 stars as \textit{negative}, 2–3 stars as \textit{neutral}, 4–5 stars as \textit{positive}).
We use the training and testing split from \citet{priban-steinberger-2021-multilingual}, see Table \ref{tab:sentiment-classification-dataset-stats}.

\begin{table}[ht!]
\begin{adjustbox}{width=0.65\linewidth,center}
\begin{tabular}{lrrrr}
\toprule
Split  & \multicolumn{1}{c}{Positive}         & \multicolumn{1}{c}{Negative}       & \multicolumn{1}{c}{Neutral}                         \\ \midrule
train & 24,573   & 23,840  & 24,691    \\
test  & 6,324    & 5,876  & 6,077       \\  \hdashline
total & 30,897   & 29,716  & 30,768     \\ 
       \bottomrule            
\end{tabular}
\end{adjustbox}
\caption{Statistics of the CSFD dataset.}
\label{tab:sentiment-classification-dataset-stats}
\end{table}

\par For the additional pre-training (see Section \ref{sec:additional-pre-training}), we downloaded 4.2M movie reviews (i.e. 1.8 GB of plain text) from the Czech movie database\footref{foot:github}. From the downloaded reviews, we removed all reviews present in the annotated CSFD dataset.

\subsection{Data for Aspect-Based Sentiment Analysis}
\label{sec:data-absa}
For the ABSA tasks, we use the Czech dataset \citep{hercig2016unsupervised} from a restaurant domain which we convert into the SemEval 2016 competition \citep{pontiki-etal-2016-semeval} format to align with the ABSA tasks addressed in this paper. The dataset consists of 2,149 Czech restaurant reviews, which we split into the training and testing parts in a 75:25 ratio. The label distribution of the modified\footnote{The dataset was converted into the SemEval 2016 competition \citep{pontiki-etal-2016-semeval}.} ABSA dataset \citep{hercig2016unsupervised} is shown in Table \ref{tab:absa-dataset-stats}, along with the number of sentiment labels for aspect categories used in the APD and TASD tasks\footnote{Because one review can contain multiple aspect categories, the number of sentiment labels does not sum up to the number of given sentences in Table \ref{tab:absa-dataset-stats}.}.

\begin{table}[ht!]
    \centering
    \begin{adjustbox}{width=0.85\linewidth,center}
    \begin{tabular}{lrrrr}
        \toprule
        Split & Sentences & Positive & Negative & Neutral \\ \midrule
        train & 1,612      &  1,231  & 1,197 & 336 \\
        test  & 537       & 420    & 426 & 61 \\  \hdashline
        total & 2,149 & 1,651  & 1,623 & 397 \\ 
        \bottomrule
    \end{tabular}
    \end{adjustbox}
    \caption{Statistics of the Czech ABSA dataset.}
    \label{tab:absa-dataset-stats}
\end{table}

\par For the additional pre-training, we scraped 2.4M reviews of Czech restaurants from Google Maps\footnote{\url{https://www.google.com/maps}}, resulting in 330 MB of plain text. As restaurant reviews are shorter, the size is smaller than downloaded movie reviews. This resulted in 330 MB of plain text, a much smaller size compared to the downloaded movie reviews due to the shorter length of restaurant reviews. We removed all reviews present in the annotated ABSA dataset.

\subsection{Aspect-Based Sentiment Tasks Definition}
\label{sec:absa-definition}
\par
Given the complexity and possible confusion in naming the aspect-based tasks we deal with in this paper, we briefly describe the tasks. As mentioned, in the ABSA tasks, we aim at the Czech restaurant reviews domain. 

\par
The ACD task aims to identify all \textit{E\#A} aspect categories towards which an opinion is expressed in a given sentence. The \textit{E\#A} represents a pair of one entity \textit{E} (i.e. Ambience, Drinks, Food, Location, Restaurant and Service), and one attribute/aspect \textit{A} (i.e. General, Miscellaneous, Prices, Quality, Style-Options). There are 14 predefined pairs of \textit{E\#A}, for example, \textit{FOOD\#PRICES}. Other than the predefined pair combinations are not allowed.

\par
The ATE aims to extract the aspect term, i.e. the linguistic expression used in the given text that represents the entity \textit{E} of each \textit{E\#A} pair. The aspect term does not have to be mentioned directly, for example, in the review: \textit{\say{Expensive but delicious}}, the entity \textit{E} is \textit{Food}, but the aspect term is not present in the text. In such cases, the \textit{NULL} value is assigned. The ACTE task focuses on extracting the aspect term and aspect category simultaneously.

\par
The APD task's goal is to assign one of the three polarity labels (\textit{positive}, \textit{negative}, \textit{neutral}) for all already identified (aspect category, aspect term) pairs in a given text. See Figure \ref{fig:absa-example} for an example.

\par
In the TASD task, the goal is to identify all (aspect category, aspect term, sentiment polarity) triplets simultaneously, which makes this task the most difficult task we solve.

% ============================== Models =====================================

\section{Models \& Approaches}
We use pre-trained Transformer-based models as backbones for our experiments. We propose a method for solving multiple ABSA tasks concurrently with sequence-to-sequence models\footnote{Also known as \textit{text-to-text} models.}, which process text (sequence) as input and produce text (sequence) as output. We employ this approach for the ACD, ATE, ACTE and TASD subtasks. To the best of our knowledge, there are no Czech monolingual sequence-to-sequence models. Therefore, we use the large \textbf{mT5} \citep{xue2020-mt5} and large \textbf{mBART} \citep{tang-etal-2021-multilingual} models, which are multilingual versions of the English T5 \citep{raffel2020exploring-t5} and BART \citep{lewis2019bart} models, respectively. 
\par
We do not use these models for the APD task as they lack prior information about the aspect term and category, which they predict along with the sentiment. The APD task assumes that the model already knows the gold data for the aspect term and category, so we would have to modify the input and output format for the APD task to make a fair comparison. Changing the output format would also be required for the SC task.

\par
Since we focus solely on the Czech language, we also wanted to evaluate Czech monolingual models. As stated above, there are no monolingual Czech sequence-to-sequence models, but only classical Czech monolingual BERT-like models such as \textbf{Czert} \citep{sido2021czert}, \textbf{RobeCzech} \citep{straka2021robeczech} or \textbf{FERNET} \citep{FERNET-2021}. Unfortunately, these models are unsuitable for our proposed approach, so we use them only for the APD and SC tasks. These models consist only of the encoder part of the Transformer architecture.

\subsection{Sequence-to-Sequence Models}
We employ the multilingual sequence-to-sequence models (mT5, mBART) to solve several ABSA tasks at once. These models consist of two parts of the Transformer architecture: the \textit{encoder} and the \textit{decoder}. Given the input sequence $x$, the encoder transforms it into a contextualized sequence $\vec{e}$. The decoder then models the conditional probability distribution of the target sequence $y$ given the encoded input $\vec{e}$ as $P_\vec{\Theta}(y|\vec{e})$, where $\vec{\Theta}$ are the parameters of the model. At each step, $i$, the decoder output $y_i$ is computed based on the previous outputs $y_0, \dots, y_{i-1}$ and the encoded input $\vec{e}$. During fine-tuning, we update all model parameters.

\subsubsection{Traditional Fine-Tuning}
Because the output of sequence-to-sequence models is text, we have to convert our discrete ABSA labels to the textual format inspired by \citet{zhang-etal-2021-aspect-sentiment}. For each example in the ABSA dataset, we construct the label as \quotes{\textit{$c$ is $P_p(p)$, given the expression: $a$}}, where $c$ is the aspect category, $a$ the aspect term and $P_p(p)$ a mapping function that maps the sentiment polarity $p$ as
\begin{equation}
    P_p(p) = 
    \begin{cases}
        \textit{great} & \text{if $p$ is \textit{positive},}\\
        \textit{ok} & \text{if $p$ is \textit{neutral},}\\
        \textit{bad} & \text{if $p$ is \textit{negative}.}\\
    \end{cases}
\end{equation}
For example, given the review: \quotes{\textit{The steak was very tasty}} the following label is generated: \quotes{\textit{Food quality} is \textit{great}, given the expression: \textit{steak}}. If an example has multiple annotation triplets\footnote{Each review can have multiple aspect categories and aspect terms, thus multiple triplet annotations.}, we concatenate the labels with semicolons.

\par
\begin{figure}
    \centering
    \includegraphics[width=7.7cm]{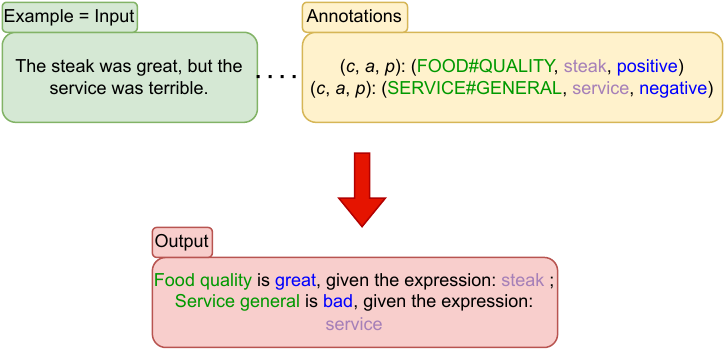}
    \caption{Example of the input and output construction for the T5 model with traditional fine-tuning.}
    \label{fig:t5traditional}
\end{figure}

 In this scenario, the model's input is the text (review), and the expected output is the textual label. The model's parameters are optimized to produce textual label in the desired format. Figure \ref{fig:t5traditional} shows an example of creating the input and target for the mT5 model with traditional fine-tuning.

\subsubsection{Prompt-Based Fine-Tuning}
For the prompt-based method, we expand the input review $x$ with a template $t$ to create a final input $x'$: $x'=x+|+t$. The template has the same form as the label in the traditional fine-tuning method. The number of transformed triplets in the prompt corresponds to the number of triplets provided for one example. We design the prompt for the mT5 and mBART models differently because their training objectives differ.

\par
\begin{figure}[ht!]
    \centering
    \includegraphics[scale=0.71]{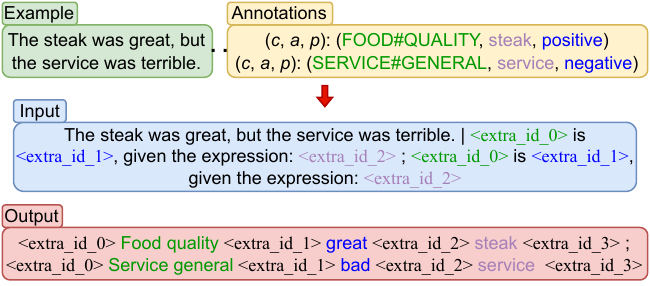}
    \caption{Example of the input and output construction or the T5 model with prompting.}
    \label{fig:t5prompt}
\end{figure}

 The mT5 model aims to reconstruct randomly selected continuous spans of input text that are masked by sentinel tokens \texttt{<extra\_id\_$id$>} during pre-training. Here, \textit{id} refers to the ID of the sentinel token, which starts from zero and increments by one. The model replaces non-masked spans of text with sentinel tokens. In our method, we replace the aspect category with \texttt{<extra\_id\_0>}, the sentiment polarity with \texttt{<extra\_id\_1>}, and the aspect term with \texttt{<extra\_id\_2>} to create the final input, which is inspired by work in \citep{gao-etal-2022-lego}. The output of the mT5 model consists of the aspect category, sentiment polarity and aspect term separated by sentinel tokens. Figure \ref{fig:t5prompt} shows an example of creating the input and target for the mT5 model with prompting.

\par
Unlike T5, the BART model reconstructs the entire input text rather than just masked spans. Furthermore, the BART model utilizes the \texttt{<mask>} token instead of sentinel tokens.

\subsubsection{Task Predictions}
As mentioned earlier, we use sequence-to-sequence models to solve multiple ABSA tasks simultaneously, namely the ACD, ATE, ACTE and TASD. Each task aims to predict different components of the annotation triplet (aspect category, aspect term, sentiment polarity). 
We generate one output for all tasks and use only the relevant part of the output for each task while discarding the rest. We can extract the relevant part for each task because the model is trained to generate output in the expected format. For instance, we extract only the aspect term from the generated output in the ATE task. For the ATE task, we consider only distinct targets and discard \textit{NULL} targets for the evaluation. For the ACD, ACTE and TASD tasks, we ignore duplicate occurrences of the predicted targets (e.g. aspect category for the ACD task).

\subsection{Sentiment Polarity Classification Models}
We use Czech BERT-like (encoder-based) models (i.e. Czert, RobeCzech, FERNET) to classify the sentiment polarity. These models convert an input sequence $x={w_1,\dots, w_k}$ of $k$ tokens into a sequence of hidden vectors $\vec{h} = \vec{h}_0,\vec{h}_1, \dots,\vec{h}_k$. For the APD task, we create $n$ input-target pairs for each example, where $n$ is the number of annotation triplets for that example.

\subsubsection{Traditional Fine-Tuning}
We employ a linear layer on top of the model to make a prediction. It computes the probability of a label $y$ from a label space $\mathcal{Y}\in\{positive, negative, neutral\}$ for the input $x_i$ as
\begin{equation}
    P_\vec{\Theta}(y|x_i)={\text{softmax}}(\vec{W}\vec{h}_{\text{\texttt{[CLS]}}} + b),
\end{equation}
where $\vec{\Theta}$ denotes all the parameters to be fine-tuned, including task-specific ones ($\vec{W}$ and $b$). The hidden vector $\vec{h}_{\text{\texttt{[CLS]}}}$ represents the artificial classification \texttt{[CLS]} token corresponding to the first hidden vector of the input sequence, i.e. $\vec{h}_{\text{\texttt{[CLS]}}}=\vec{h}_0$, and represents the entire input sequence. 

\par
In the case of the ABSA dataset and the SC of the aspect term and category, we append the aspect term and category to the beginning of the input so that the model has the knowledge of the specific tuple by which to make predictions.

\subsubsection{Prompt-Based Fine-Tuning}
For prompt-based fine-tuning, we exploit the fact that the models were pre-trained by the masked language modelling task \citep{devlin-etal-2019-bert}. We use the language modelling property of the model to generate a token that represents the polarity label.

\begin{figure}
    \centering
    \includegraphics[scale=0.67]{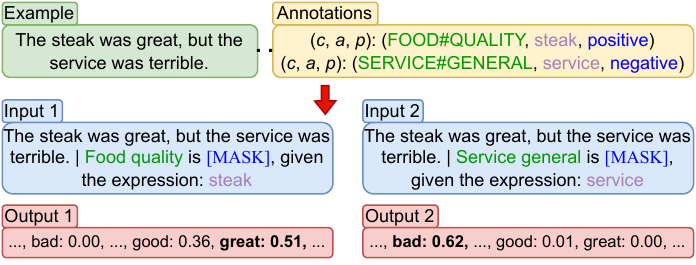}
    \caption{Example of the input and output construction for the classification model using prompting.}
    \label{fig:class_prompt}
\end{figure}

\par
During prompt-based fine-tuning, we create a new input $x'$ from the original input $x$ by appending a task-specific prompt. The prompt has one answer slot represented by a \texttt{[MASK]} token, which the model fills with the highest-probability token from its vocabulary for the given context. Each label from label space $\mathcal{Y}$ is mapped to a word from the model's vocabulary $\mathcal{V}$ using a mapping $\mathcal{M} = \mathcal{Y} \rightarrow \mathcal{V}$, which is for Czech defined as follows

\begin{equation}
    P_p(p) = 
    \begin{cases}
        \textit{dobrý} & \text{if $p$ is \textit{positive},}\\
        \textit{ok} & \text{if $p$ is \textit{neutral},}\\
        \textit{špatný} & \text{if $p$ is \textit{negative}.}\\
    \end{cases}
    \label{eqn:sentiment_mapping}
\end{equation}

Figure \ref{fig:class_prompt} shows an example of input construction with desired outputs for the ABSA task.

\par
We trim the original input of long reviews before appending the prompt to ensure that the new input $x'$ fits into the model. We use different prompts for each dataset. For the CSFD dataset, we use the prompt \quotes{\textit{Je to }\texttt{[MASK]} \textit{film}} (\quotes{\textit{It is a }\texttt{[MASK]} \textit{movie}} in English). For the ABSA dataset, the prompt is structured as \quotes{$c$ \textit{je} \texttt{[MASK]}, \textit{dáno výrazem:} $a$} (\quotes{$c$ \textit{is} \texttt{[MASK]}, \textit{given the expression:} $a$} in English), where $c$ is the aspect category (translated to Czech) and $a$ is the aspect term.

% ============================== Experiments \& Results =====================================
\section{Experiments \& Results}
In our experiments, we fine-tune the sequence-to-sequence models (mBART, mT5) for the ATE, ACD, ACTE, and TASD tasks on the entire ABSA dataset using both traditional and prompt-based fine-tuning approaches and we report the results as micro F1 scores. The BERT-like (encoder-based) models (Czert, RobeCzech and FERNET) are fine-tuned for the APD and SC tasks and report results as accuracy. For these tasks, we further experiment with zero-shot and few-shot scenarios, as well as additional pre-training of the Czech models.

\par
To ensure the reliability of our results, we perform each experiment five times with different random seed initialization and report the average scores along with a 95\% confidence interval. We provide the training details in Appendix \ref{sec:transformer-appendix}.

\subsection{Few-Shot and Zero-Shot Setting}
In the few-shot setting, we fine-tune the models on the first $n$ examples of the training data using a fixed training set to ensure a fair comparison between models, as recommended by \citet{schick-schutze-2021-just}. 
In the zero-shot setting, models are evaluated on the test set without any fine-tuning.

\subsection{Additional Pre-Training}
\label{sec:additional-pre-training}
For the APD and SC tasks, we were interested in whether additional pre-training in the task domain helps to improve results. Therefore, we further pre-train the three Czech models (Czert, RobeCzech and FERNET) with the masked language modelling task on restaurant reviews and movie reviews for the APD and SC tasks, respectively. See Appendix \ref{sec:appendix-additional-pre-training} for details.

\subsection{Results for Aspect-Based Sentiment}
Table \ref{tab:absaseq2seq} shows the results achieved by the sequence-to-sequence models. The prompting approach (PT-FT) significantly enhances the performance of both models. Without prompting, i.e. with the traditional fine-tuning (TR-FT), mBART performs better than mT5. However, with prompting, mT5 performs better than or similar to mBART.

\begin{table}[ht!]
\begin{adjustbox}{width=1\linewidth,center}
    \begin{tabular}{@{}lccccc@{}}
        \toprule
        \multirow{2}{*}{\textbf{Model}} & \multirow{2}{*}{\textbf{Approach}} & \multicolumn{4}{c}{\textbf{Task}}                                                \\ \cmidrule(lr){3-6} 
                         &      & ACD               & ATE               & ACTE  & TASD                     \\ \midrule
        \multirow{2}{*}{mT5}  & TR-FT                 & 75.5$^{\pm1.8}$ & 66.5$^{\pm2.5}$ & 56.4$^{\pm1.0}$ & 48.0$^{\pm1.0}$ \\
        
        & PT-FT  & \textbf{\underline{85.5}}$^{\pm1.2}$ & \textbf{\underline{84.8}}$^{\pm1.6}$ & \textbf{\underline{75.0}}$^{\pm1.9}$ & \textbf{\underline{67.3}}$^{\pm1.7}$ \\ \cdashlinelr{2-6}
        \multirow{2}{*}{mBART}        & TR-FT           & 78.7$^{\pm1.6}$ & 78.9$^{\pm1.3}$ & 67.2$^{\pm1.4}$ & 57.5$^{\pm1.7}$ \\
        
      & PT-FT  & \underline{83.3}$^{\pm0.7}$ & \underline{83.4}$^{\pm0.6}$ & \underline{71.9}$^{\pm1.6}$ & \underline{61.7}$^{\pm0.7}$ \\ \bottomrule
    \end{tabular}
    \end{adjustbox}
    \caption{Results of the sequence-to-sequence models as micro F1 scores on different ABSA tasks with traditional fine-tuning (TR-FT) and prompt-based fine-tuning (PT-FT). The best results for each task are in \textbf{bold}.
    \underline{Underlined} results indicate significantly better performance between the two fine-tuning styles for a given model and task.
    }
    \label{tab:absaseq2seq}
    
\end{table}

\begin{table*}[ht!]
\begin{adjustbox}{width=0.99\linewidth,center}
\begin{tabular}{lllllll} \toprule
     & \multicolumn{2}{c}{\textbf{Czert}}                                & \multicolumn{2}{c}{\textbf{RobeCzech}}                            & \multicolumn{2}{c}{\textbf{FERNET}}                \\ \cmidrule(lr){2-3} \cmidrule(lr){4-5} \cmidrule(lr){6-7}
         & \multicolumn{1}{c}{TR-FT}          & \multicolumn{1}{c}{PT-FT}          & \multicolumn{1}{c}{TR-FT}          & \multicolumn{1}{c}{PT-FT}          & \multicolumn{1}{c}{TR-FT}          & \multicolumn{1}{c}{PT-FT}          \\
     & \multicolumn{1}{c}{original/pre-train} & \multicolumn{1}{c}{original/pre-train} & \multicolumn{1}{c}{original/pre-train} & \multicolumn{1}{c}{original/pre-train} & \multicolumn{1}{c}{original/pre-train} & \multicolumn{1}{c}{original/pre-train} \\  \hline %\midrule          
    
\multicolumn{7}{l}{\textit{Zero-shot}}                                                                                                                                                                           \\
  & \underline{47.1}$^{\pm0.7}$/\underline{42.4}$^{\pm6.4}$ & 5.3$^{\pm0.0}$/5.3$^{\pm0.0}$   & \textbf{\underline{47.4}}$^{\pm2.5}$/\underline{42.3}$^{\pm3.5}$ & 8.4$^{\pm0.0}$/3.8$^{\pm0.0}$   & \underline{43.6}$^{\pm2.4}$/\underline{43.2}$^{\pm3.3}$ & 0.8$^{\pm0.0}$/3.2$^{\pm0.0}$   \\ \cdashlinelr{2-7}
\multicolumn{7}{l}{\textit{Fine-tuning (few-shot)}}                                                                                                                                                              \\
10   & 46.0$^{\pm1.9}$/55.5$^{\pm3.9}$ & \underline{67.6}$^{\pm3.6}$/\underline{77.5}$^{\pm5.0}$ & 47.5$^{\pm3.0}$/65.5$^{\pm6.9}$ & \underline{77.3}$^{\pm3.4}$/\textbf{\underline{81.9}}$^{\pm1.4}$ & 48.8$^{\pm2.0}$/66.3$^{\pm4.0}$ & \underline{77.6}$^{\pm6.5}$/\underline{76.9}$^{\pm3.7}$ \\
20   & 54.6$^{\pm6.0}$/76.5$^{\pm5.4}$ & \underline{74.3}$^{\pm1.3}$/80.4$^{\pm1.1}$ & 59.7$^{\pm2.0}$/63.4$^{\pm7.3}$ & \underline{78.5}$^{\pm2.0}$/\textbf{\underline{82.8}}$^{\pm1.1}$ & 62.6$^{\pm1.6}$/79.4$^{\pm4.2}$ & \underline{72.7}$^{\pm3.0}$/78.5$^{\pm2.1}$ \\
50   & 66.0$^{\pm4.6}$/83.4$^{\pm2.3}$ & \underline{75.2}$^{\pm2.0}$/80.9$^{\pm2.6}$ & 75.3$^{\pm2.5}$/\textbf{86.7}$^{\pm1.4}$ & \underline{83.0}$^{\pm1.7}$/85.7$^{\pm0.6}$ & 71.9$^{\pm2.1}$/83.7$^{\pm3.3}$ & \underline{84.5}$^{\pm1.4}$/86.6$^{\pm1.4}$ \\
100  & 66.6$^{\pm3.0}$/80.4$^{\pm1.4}$ & \underline{75.9}$^{\pm0.7}$/81.2$^{\pm1.8}$ & 76.3$^{\pm6.9}$/84.3$^{\pm1.6}$ & 83.3$^{\pm1.3}$/\textbf{85.5}$^{\pm1.0}$ & 71.6$^{\pm2.7}$/82.5$^{\pm2.1}$ & \underline{84.1}$^{\pm1.6}$/85.1$^{\pm1.7}$ \\
500  & 81.4$^{\pm2.1}$/84.1$^{\pm1.4}$ & 82.6$^{\pm1.0}$/84.3$^{\pm0.9}$ & 84.0$^{\pm1.4}$/\underline{86.6}$^{\pm0.3}$ & 85.6$^{\pm1.8}$/85.3$^{\pm0.8}$ & 84.5$^{\pm1.1}$/83.8$^{\pm0.5}$ & 84.2$^{\pm1.1}$/\textbf{\underline{86.7}}$^{\pm1.6}$ \\
1,000 & 82.0$^{\pm1.1}$/83.4$^{\pm1.6}$ & 82.7$^{\pm1.0}$/83.2$^{\pm1.5}$ & 83.1$^{\pm2.7}$/\textbf{87.4}$^{\pm2.1}$ & 85.3$^{\pm1.7}$/87.2$^{\pm1.5}$ & 84.6$^{\pm0.8}$/87.0$^{\pm1.1}$ & \underline{85.9}$^{\pm0.5}$/85.9$^{\pm0.7}$ \\ \cdashlinelr{2-7}
\multicolumn{7}{l}{\textit{Fine-tuning (full)}}                                                                                                                                                                  \\
   & 83.2$^{\pm1.4}$/85.0$^{\pm1.1}$ & 84.2$^{\pm1.1}$/87.0$^{\pm1.3}$ & 85.2$^{\pm1.6}$/88.4$^{\pm0.9}$ & 87.3$^{\pm1.4}$/\textbf{88.7}$^{\pm1.0}$ & 86.0$^{\pm0.4}$/88.4$^{\pm0.7}$ & 87.5$^{\pm1.2}$/88.5$^{\pm0.7}$ \\ \bottomrule
\end{tabular}
\end{adjustbox}
\caption{Results for the ABSA dataset on APD task as accuracy with prompt-based fine-tuning (PT-FT) and traditional fine-tuning (TR-FT) approaches.
The best results for a given configuration are in \textbf{bold}. \underline{Underlined} results indicate significantly better performance between the two fine-tuning styles for a given model (both original and with additional pre-training) and the number of training examples.
}
\label{tab:absa}
\end{table*}

\par
The best results are achieved on the ACD task. For this task, there is a predefined set of categories. In contrast, the ATE task poses a greater challenge, as the extracted term can be an arbitrarily long sequence of different words, making this task more difficult. The ACTE is even more challenging since the model has to simultaneously predict the aspect term and category. The TASD task is the most difficult of the solved tasks because the model must predict the aspect term, aspect category and sentiment polarity simultaneously. 
% Because our study is the first one that focuses on these tasks in Czech, there are no results from other studies which we could compare.
Since our study is the first to focus on these tasks in the Czech language, we lack a basis for comparison with other studies.

Table \ref{tab:absa} shows the results of the APD task. Traditional fine-tuning performs significantly better than prompting in the zero-shot setting. Prompting outperforms traditional fine-tuning when using a small number of examples for training. In the rest of the results, both fine-tuning approaches perform similarly. The domain pre-training improves the results of all models, especially for traditional fine-tuning.

\subsection{Sentiment Classification Results}
Table \ref{tab:csfd} shows the sentiment classification results on the CSFD dataset, along with the current SotA results. In the zero-shot setting, the traditional fine-tuning approach (TR-FT) yields random results around 35–38\%. This is expected because the linear layer\footnote{The layer always returns one of three possible labels, thus if the dataset is perfectly balanced, the random (and also lowest) expected accuracy is $33.\overline{3}$\%.} on top of the model is not trained and the CSFD dataset contains three roughly balanced classes. On the other hand, the zero-shot scenario with the prompt-based approach\footnote{In this case, the model can predict any word from the model vocabulary $\mathcal{V}$; therefore, the potential lowest expected random accuracy is close to zero ($1/|\mathcal{V}|$).} (PT-FT) combined with the additional domain pre-training provides significantly better results for Czert and FERNET models, achieving 48.2\% and 59.2\%, respectively.

We observed that prompting consistently outperforms traditional fine-tuning in the few-shot scenario with 10 and 20 training examples. In contrast, traditional fine-tuning yields better results when using 100, 500 and 1,000 examples. Results are comparable for both approaches when the model is trained on all examples and 50 examples.
Domain pre-training improves the results in most cases, especially when using only a small number of examples.
Notably, the FERNET model achieved the best result of 88.2\% accuracy, surpassing the current SotA by 2.8\%.

\begin{table*}[ht!]
\begin{adjustbox}{width=1\linewidth,center}
\begin{tabular}{p{0.5\columnwidth}lllllll} \toprule
     & \multicolumn{2}{c}{\textbf{Czert}}                                & \multicolumn{2}{c}{\textbf{RobeCzech}}                            & \multicolumn{2}{c}{\textbf{FERNET}}                                         \\ \cmidrule(lr){2-3} \cmidrule(lr){4-5} \cmidrule(lr){6-7}
         & \multicolumn{1}{c}{TR-FT}          & \multicolumn{1}{c}{PT-FT}          & \multicolumn{1}{c}{TR-FT}          & \multicolumn{1}{c}{PT-FT}          & \multicolumn{1}{c}{TR-FT}          & \multicolumn{1}{c}{PT-FT}          \\
     & \multicolumn{1}{c}{original/pre-train} & \multicolumn{1}{c}{original/pre-train} & \multicolumn{1}{c}{original/pre-train} & \multicolumn{1}{c}{original/pre-train} & \multicolumn{1}{c}{original/pre-train} & \multicolumn{1}{c}{original/pre-train} \\  \hline %\midrule
\multicolumn{7}{l}{\textit{Zero-shot}}                                                                                                                                                                           \\
    & \underline{35.0}$^{\pm0.7}$/35.7$^{\pm2.2}$ & 11.8$^{\pm0.0}$/\underline{48.2}$^{\pm0.0}$ & \underline{36.3}$^{\pm2.9}$/\underline{35.7}$^{\pm5.0}$ & 12.7$^{\pm0.0}$/8.9$^{\pm0.0}$  & \underline{38.2}$^{\pm1.1}$/36.8$^{\pm4.0}$ & 5.8$^{\pm0.0}$/\textbf{\underline{59.2}}$^{\pm0.0}$  \\ \cdashlinelr{2-7}
\multicolumn{7}{l}{\textit{Fine-tuning (few-shot)}}                                                                                                                                                              \\ 
10   & 43.4$^{\pm1.9}$/54.6$^{\pm2.0}$ & \underline{50.3}$^{\pm0.6}$/\underline{60.4}$^{\pm0.8}$ & 46.2$^{\pm3.0}$/61.3$^{\pm0.4}$ & \underline{54.6}$^{\pm1.4}$/\textbf{62.4}$^{\pm1.5}$ & 48.8$^{\pm2.5}$/55.1$^{\pm3.3}$ & \underline{56.4}$^{\pm0.6}$/\underline{61.5}$^{\pm0.5}$ \\
20   & 47.4$^{\pm3.1}$/60.9$^{\pm3.6}$ & \underline{51.5}$^{\pm0.3}$/65.2$^{\pm1.0}$ & 48.4$^{\pm3.3}$/65.4$^{\pm4.1}$ & \underline{56.0}$^{\pm0.9}$/\textbf{\underline{72.3}}$^{\pm0.8}$ & 57.8$^{\pm2.9}$/62.6$^{\pm2.8}$ & \underline{62.6}$^{\pm1.7}$/\underline{67.5}$^{\pm0.3}$ \\
50   & 57.1$^{\pm3.6}$/71.0$^{\pm1.2}$ & 58.7$^{\pm0.8}$/70.9$^{\pm0.7}$ & 56.7$^{\pm4.7}$/\textbf{\underline{78.5}}$^{\pm0.9}$ & 60.3$^{\pm2.2}$/77.1$^{\pm0.4}$ & 66.6$^{\pm2.4}$/74.7$^{\pm4.2}$ & 67.4$^{\pm1.8}$/75.7$^{\pm3.9}$ \\
100  & \underline{64.3}$^{\pm0.8}$/\underline{73.9}$^{\pm0.7}$ & 61.6$^{\pm0.6}$/72.8$^{\pm0.2}$ & \underline{69.8}$^{\pm1.1}$/\textbf{\underline{80.1}}$^{\pm0.3}$ & 67.7$^{\pm0.9}$/78.6$^{\pm0.2}$ & \underline{74.1}$^{\pm0.4}$/79.8$^{\pm1.0}$ & 72.1$^{\pm0.4}$/78.2$^{\pm1.2}$ \\
500  & \underline{70.7}$^{\pm0.2}$/75.7$^{\pm1.0}$ & 69.2$^{\pm0.4}$/75.8$^{\pm0.3}$ & 74.3$^{\pm0.7}$/\underline{82.2}$^{\pm0.2}$ & 73.9$^{\pm0.5}$/81.1$^{\pm0.2}$ & \underline{77.3}$^{\pm0.3}$/\textbf{82.5}$^{\pm0.5}$ & 76.4$^{\pm0.1}$/81.8$^{\pm0.4}$ \\
1,000 & \underline{72.7}$^{\pm0.1}$/76.6$^{\pm0.1}$ & 71.2$^{\pm0.2}$/76.1$^{\pm0.9}$ & 76.2$^{\pm0.8}$/\underline{82.7}$^{\pm0.2}$ & 75.7$^{\pm0.7}$/82.3$^{\pm0.1}$ & \underline{78.4}$^{\pm0.3}$/\textbf{83.0}$^{\pm0.8}$ & 77.6$^{\pm0.3}$/82.3$^{\pm0.4}$ \\  \cdashlinelr{2-7}
\multicolumn{7}{l}{\textit{Fine-tuning (full)}}                                                                                                                                                                  \\
    & 85.3$^{\pm0.1}$/86.5$^{\pm0.1}$ & 85.3$^{\pm0.1}$/86.3$^{\pm0.1}$ & 87.1$^{\pm0.0}$/88.0$^{\pm0.1}$ & 87.0$^{\pm0.3}$/87.9$^{\pm0.2}$ & 87.3$^{\pm0.1}$/\textbf{88.2}$^{\pm0.1}$ & 87.2$^{\pm0.2}$/87.7$^{\pm0.7}$ \\  \cdashlinelr{1-7}
\citet{priban-steinberger-2021-multilingual}
  & 84.8$^{\pm0.1}$/ \phantom{*} -  & \multicolumn{1}{c}{-} & \multicolumn{1}{c}{-} & \multicolumn{1}{c}{-} & \multicolumn{1}{c}{-}  & \\
    \citet{FERNET-2021}
   & \multicolumn{1}{c}{-}  & \multicolumn{1}{c}{-}  & 85.0$^{\pm0.4}$/ \phantom{*} -  & \multicolumn{1}{c}{-}   & 85.4$^{\pm0.3}$/ \phantom{*} -   & \multicolumn{1}{c}{-}  \\

    \bottomrule
\end{tabular}
\end{adjustbox}
 \caption{Sentiment classification results for the CSFD dataset as accuracy with prompt-based fine-tuning (PT-FT) and traditional fine-tuning (TR-FT) approaches.
 The best results for a given configuration are in \textbf{bold}. \underline{Underlined} results indicate significantly better performance between the two fine-tuning styles for a given model (both original and with additional pre-training) and the number of training examples.
 }
\label{tab:csfd}
\end{table*}

\subsection{Discussion}
The prompt designed for the APD task might be more suitable than the prompt for the SC task, which may explain why prompting is worse only in one case than traditional fine-tuning outside of the zero-shot setting, while traditional fine-tuning outperforms prompting more often in the SC task.

\par
The reason why the sequence-to-sequence models perform better with prompting than with traditional fine-tuning may be that the prompting matches these models’ pre-training objectives closely. Additionally, these models possess some prior information about the number of sentiment triplets they should generate in the prompt, which the traditional fine-tuned models do not.

\par
Our research indicates that the sequence-to-sequence models have no problems generating the output in the required format, which is crucial to extract the targets. However, when using traditional fine-tuning, the mT5 model occasionally generates repeated transformed triplets and lacks diversity in its output more frequently than the mBART model, which may explain why the mBART model outperforms the mT5 model with traditional fine-tuning.

\par
We observe a common trend in results for SC and APD tasks, whereby the prompting approach with a smaller number of training examples outperforms the traditional fine-tuning, which is consistent with conclusions from \citet{gao-etal-2021-making}.

\par
For prompting in few-shot and zero-shot scenarios, a mapping function that maps one sentiment to multiple words instead of one specific word would likely lead to better results, which can be explored in future work.

% ============================== Conclusion =====================================
\section{Conclusion}
In this work, we introduced a sequence-to-sequence method that solves multiple ABSA tasks simultaneously and can be used with both traditional fine-tuning and prompting. Experiments on the Czech dataset show that prompting significantly improves performance. Furthermore, we proposed a method for sentiment classification that can also be used with prompting and traditional fine-tuning. We evaluate this method on two Czech datasets with three monolingual Czech models and demonstrate the effectiveness of prompting for few-shot fine-tuning, where prompting consistently outperforms the traditional approach. Finally, we show that pre-training on the domain data significantly enhances the results, especially in a zero-shot scenario.

\section*{Acknowledgments}
This work has been partly supported by grant No. SGS-2022-016 Advanced methods of data processing and analysis.
Computational resources were provided by the e-INFRA CZ project (ID:90140), supported by the Ministry of Education, Youth and Sports of the Czech Republic.

\bibliographystyle{acl_natbib}
\bibliography{anthology,ranlp2023}

\appendix

\section{Appendix}
\label{sec:appendix}
\subsection{Hyper-parameters \& Training Details}
\label{sec:transformer-appendix}

\par
We train the models with different hyper-parameters and select the best-performing model based on the performance on the validation data, including the number of epochs. The CSFD dataset is already split into training and validation data. For the ABSA dataset, we use 10\% of the training data as validation data. The final experiments are conducted on all training data and evaluated on the test data.
\par
We use a batch size of 64 and train the sequence-to-sequence models for up to 35 epochs. For the mT5 model, we search for a learning rate from \{1e-4, 3e-4\}, while for the mBART model, we search for a learning rate from \{5e-5, 1e-5\}. We use greedy search for simplicity because experiments with a beam search with beam sizes 3 and 5 lead to similar performance.
\par
For the models for sentiment polarity classification, we search for a learning rate from \{5e-5, 1e-5\}. We use up to 10 epochs and a batch size of 16 for the CSFD dataset and up to 50 epochs and a batch size of 64 for the ABSA dataset.
\par
We optimize the cross-entropy loss for all the models. All the models have the maximum input sequence length limited to 512 tokens. We use the AdaFactor \citep{adafactor} optimizer for the mT5 model and AdamW \citep{adamw} for the rest of the models. We keep the default dropout value for all the models, which is 0.1.
\par For text generation with the sequence-to-sequence models, we use the \textit{AutoModelForSeq2SeqLM} class with greedy search decoding from the HuggingFace library\footnote{\url{https://huggingface.co}}. We tried different configurations of the beam search decoding algorithm \citep{freitag-al-onaizan-2017-beam}, but it provides the same results as the greedy search algorithm, so we employ the greedy search algorithm for simplicity.

\subsection{Details of Additional Pre-Training}
\label{sec:appendix-additional-pre-training}
The additional pre-training of Czert, RobeCzech and FERNET models on data from a specific task domain (restaurant reviews and movie reviews) is performed with the masked language modelling task \citep{devlin-etal-2019-bert}. The pre-training process was carried out with a batch size of 512 and a maximum input sequence length of 512 for all models. We optimize the models with the cross-entropy loss function and AdamW \citep{adamw} optimizer for 20K batches (steps). We use a learning rate of 5e-5 with linear decay. The word masking probability is set to 15\%.

\end{document}